\documentclass{article} 
\usepackage{nips15submit_e,times}
\usepackage{hyperref}
\usepackage{url}

\usepackage{amsmath}
\usepackage{amsfonts}
\usepackage{amsthm}
\usepackage{amssymb}
\usepackage{fixmath}
\usepackage{mathtools}

\usepackage[normalem]{ulem}

\usepackage[numbers]{natbib}
\renewcommand{\vec}[1]{\mathbold{#1}}
\newcommand{\mat}[1]{\mathbold{#1}}
\newcommand{\tens}[1]{\boldsymbol{\mathcal{#1}}}
\newcommand{\tensel}[1]{\mathcal{#1}}
\DeclareMathOperator{\rank}{r}

\usepackage{graphicx,xcolor}

\graphicspath{{figures/}}
\newcommand{\executeiffilenewer}[3]{%
  \ifnum\pdfstrcmp{\pdffilemoddate{#1}}%
  {\pdffilemoddate{#2}} > 0 {\immediate\write18{#3}}\fi}

\usepackage{pgfplots}
\usetikzlibrary{plotmarks}
\usetikzlibrary{backgrounds}
\newlength\figureheight
\newlength\figurewidth
\pgfplotsset{
  /tikz/font = \scriptsize,
  compat = newest,
  tick align = outside,
  xminorticks = false,
  yminorticks = false,
  xlabel style = {yshift = -0.1cm},
  ylabel style = {yshift = 0.1cm},
  xticklabel style = {yshift = -0.05cm},
  tick pos = left,
  every axis plot/.append style = {line width = 1pt, mark size = 0.8pt},
  axis line style = {semithick},
  legend style = {semithick},
  legend image post style={mark size=1.5pt},
  tick style = {semithick, color = black},
  label style = {font = \normalsize},
}
\tikzset{
  every picture/.append style = {tight background}
}

\title{Tensorizing Neural Networks}

\author{
Alexander Novikov$^{1,4}$  ~~~~ Dmitry Podoprikhin$^{1}$ ~~~~ Anton Osokin$^{2}$ ~~~~ Dmitry Vetrov$^{1,3}$\\
$^1$Skolkovo Institute of Science and Technology, Moscow, Russia \\
$^2$INRIA, SIERRA project-team, Paris, France  \\
$^3$National Research University Higher School of Economics, Moscow, Russia \\
$^4$Institute of Numerical Mathematics of the Russian Academy of Sciences, Moscow, Russia  \\
\texttt{novikov@bayesgroup.ru} ~~~~ \texttt{podoprikhin.dmitry@gmail.com}\\
\texttt{anton.osokin@inria.fr} ~~~~ \texttt{vetrovd@yandex.ru}
}

\nipsfinalcopy 

\begin{document}

\maketitle

\begin{abstract}
Deep neural networks currently demonstrate state-of-the-art performance in several domains.
At the same time, models of this class are very demanding in terms of computational resources. In particular, a large amount of memory is required by commonly used fully-connected layers, making it hard to use the models on low-end devices and stopping the further increase of the model size. In this paper we convert the dense weight matrices of the fully-connected layers to the Tensor Train~\cite{oseledets2011ttMain} format such that the number of parameters is reduced by a huge factor and at the same time the expressive power of the layer is preserved.
In particular, for the Very Deep VGG networks~\cite{simonyan15} we report the compression factor of the dense weight matrix of a fully-connected layer up to 200000 times leading to the compression factor of the whole network up to 7 times.
\end{abstract}

\section{Introduction \label{sec:introduction}}
Deep neural networks currently demonstrate state-of-the-art performance in many domains of large-scale machine learning, such as computer vision, speech recognition, text processing, etc.
These advances have become possible because of algorithmic advances, large amounts of available data, and modern hardware.
For example, convolutional neural networks (CNNs)~\cite{Krizhevsky2012AlexNet,simonyan15} show by a large margin superior performance on the task of image classification.
These models have thousands of nodes and millions of learnable parameters and are trained using millions of images~\cite{Russakovsky2015ImageNet} on powerful Graphics Processing Units (GPUs).

The necessity of expensive hardware and long processing time are the factors that complicate the application of such models on conventional desktops and portable devices.
Consequently, a large number of works tried to reduce both hardware requirements (e.\,g. memory demands) and running times (see Sec.~\ref{sec:related-work}).

In this paper we consider probably the most frequently used layer of the neural networks: the fully-connected layer. This layer consists in a linear transformation of a high-dimensional input signal to a high-dimensional output signal with a large dense matrix defining the transformation. For example, in modern CNNs the dimensions of the input and output signals of the fully-connected layers are of the order of thousands, bringing the number of parameters of the fully-connected layers up to millions.

We use a compact multiliniear format -- Tensor-Train (TT-format)~\cite{oseledets2011ttMain} -- to represent the dense weight matrix of the fully-connected layers using few parameters while keeping enough flexibility to perform signal transformations.
The resulting layer is compatible with the existing training algorithms for neural networks  because all the derivatives required by the back-propagation algorithm~\cite{rumelhart1986} can be computed using the properties of the TT-format. We call the resulting layer a \emph{TT-layer} and refer to a network with one or more TT-layers as \emph{TensorNet}.

We apply our method to popular network architectures proposed for several datasets of different scales: MNIST~\cite{lecun1998mnist}, CIFAR-10~\cite{krizhevsky2009Cifar}, ImageNet~\cite{Krizhevsky2012AlexNet}.
We experimentally show that the networks with the TT-layers match the performance of their uncompressed counterparts but require up to $200\,000$ times less of parameters, decreasing the size of the whole network by a factor of~$7$.

The rest of the paper is organized as follows.
We start with a review of the related work in Sec.~\ref{sec:related-work}. We introduce necessary notation and review the Tensor Train (TT) format in Sec.~\ref{sec:TT-description}. In Sec.~\ref{sec:TT-layer} we apply the TT-format to the weight matrix of a fully-connected layer and in Sec.~\ref{sec:learning} derive all the equations necessary for applying the back-propagation algorithm. In Sec.~\ref{sec:experiments} we present the experimental evaluation of our ideas followed by a discussion in Sec.~\ref{sec:discussion}.\vspace{-0.05cm}

\section{Related work \label{sec:related-work}}
With sufficient amount of training data, big models usually outperform smaller ones. However state-of-the-art neural networks reached the hardware limits both in terms the computational power and the memory.

In particular, modern networks reached the memory limit with $89\%$~\cite{simonyan15} or even $100\%$~\cite{xue2013restructuring} memory occupied by the weights of the fully-connected layers so it is not surprising that numerous attempts have been made to make the fully-connected layers more compact. One of the most straightforward approaches is to use a low-rank representation of the weight matrices. Recent studies show that the weight matrix of the fully-connected layer is highly redundant and by restricting its matrix rank it is possible to greatly reduce the number of parameters without significant drop in the predictive accuracy~\cite{Denil2013predicting, sainath2013low, xue2013restructuring}.

An alternative approach to the problem of model compression is to tie random subsets of weights using special hashing techniques~\cite{chen2015compressing}. The authors reported the compression factor of $8$ for a two-layered network on the MNIST dataset without loss of accuracy. Memory consumption can also be reduced by using lower numerical precision~\cite{asanovic1991experimental} or allowing fewer possible carefully chosen parameter values~\cite{gong2014PQcompressing}.

In our paper we generalize the low-rank ideas. Instead of searching for low-rank approximation of the weight matrix we treat it as multi-dimensional tensor and apply the Tensor Train decomposition algorithm \cite{oseledets2011ttMain}. This framework has already been successfully applied to several data-processing tasks, e.\,g.~\cite{novikov2014putting, zhang2015enabling}.

Another possible advantage of our approach is the ability to use more hidden units than was available before. A recent work \cite{Jimmy2014lowRankSNN} shows that it is possible to construct wide and shallow (i.\,e. not deep) neural networks with performance close to the state-of-the-art deep CNNs by training a shallow network on the outputs of a trained deep network. They report the improvement of performance with the increase of the layer size and used up to $30\,000$ hidden units while restricting the matrix rank of the weight matrix in order to be able to keep and to update it during the training. Restricting the TT-ranks of the weight matrix (in contrast to the matrix rank) allows to use much wider layers potentially leading to the greater expressive power of the model. We demonstrate this effect by training a very wide model ($262\,144$ hidden units) on the CIFAR-10 dataset that outperforms other non-convolutional networks.

Matrix and tensor decompositions were recently used to speed up the inference time of CNNs~\cite{Denil2014speedup,lebedev2014speeding}. While we focus on fully-connected layers, Lebedev et al.~\cite{lebedev2014speeding} used the CP-decomposition to compress a $4$-dimensional convolution kernel and then used the properties of the decomposition to speed up the inference time. This work shares the same spirit with our method and the approaches can be readily combined.

Gilboa et al. exploit the properties of the Kronecker product of matrices to perform fast matrix-by-vector multiplication~\cite{gilboa2012gp}. These matrices have the same structure as TT-matrices with unit TT-ranks.

Compared to the Tucker format~\cite{tucker66} and the canonical format~\cite{caroll70}, the TT-format is immune to the curse of dimensionality and its algorithms are robust. Compared to the Hierarchical Tucker format~\cite{hackbush09hierahical}, TT is quite similar but has simpler algorithms for basic operations.

\section{TT-format \label{sec:TT-description}}
Throughout this paper we work with arrays of different dimensionality. We refer to the one-dimensional arrays as \emph{vectors}, the two-dimensional arrays~-- \emph{matrices}, the arrays of higher dimensions~-- \emph{tensors}. Bold lower case letters (e.\,g. $\vec{a}$) denote vectors, ordinary lower case letters (e.\,g. $a(i) = a_i$)~-- vector elements, bold upper case letters (e.\,g. $\mat{A}$)~-- matrices, ordinary upper case letters (e.\,g. $A(i, j)$)~-- matrix elements, calligraphic bold upper case letters (e.\,g. $\tens{A}$)~-- for tensors and ordinary calligraphic upper case letters (e.\,g. $\tensel{A}(\vec{i}) = \tensel{A}(i_1, \ldots, i_d)$)~-- tensor elements, where $d$ is the dimensionality of the tensor $\tens{A}$.

We will call arrays \emph{explicit} to highlight cases when they are stored explicitly, i.\,e. by enumeration of all the elements.

A $d$-dimensional array (tensor)~$\tens{A}$ is said to be represented in the \emph{TT-format} \citep{oseledets2011ttMain} if for each dimension~$k=1,\ldots,d$ and for each possible value of the $k$-th dimension index~$j_k = 1, \ldots, n_k$ there exists a matrix~$\mat{G}_k[j_k]$ such that all the elements of~$\tens{A}$ can be computed as the following matrix product:
\begin{equation}
\label{TT-format}
\tensel{A}(j_1, \dots, j_d) = \mat{G}_1[j_1] \mat{G}_2[j_2] \dotsm \mat{G}_d[j_d].
\end{equation}
All the matrices~$\mat{G}_k[j_k]$ related to the same dimension~$k$ are restricted to be of the same size~$\rank_{k-1} \times \rank_k$.
The values~$\rank_0$ and $\rank_d$ equal $1$ in order to keep the matrix product~\eqref{TT-format} of size $1 \times 1$.
In what follows we refer to the representation of a tensor in the TT-format as the \emph{TT-representation} or the \emph{TT-decomposition}.
The sequence~$\left\{\rank_k\right\}_{k=0}^d$ is referred to as the \emph{TT-ranks} of the TT-representation of~$\tens{A}$ (or the \emph{ranks} for short), its maximum~-- as the \emph{maximal TT-rank} of the TT-representation of~$\tens{A}$:~$\rank = \max_{k = 0, \dots, d} \rank_k$.
The collections of the matrices $\left ( \mat{G}_k[j_k] \right)_{j_k = 1}^{n_k}$ corresponding to the same dimension (technically, 3-dimensional arrays $\tens{G}_k$) are called the \emph{cores}.

\citet[Th.~2.1]{oseledets2011ttMain} shows that for an arbitrary tensor~$\tens{A}$ a TT-representation exists but is not unique. The ranks among different TT-representations can vary and it's natural to seek a representation with the lowest ranks.

We use the symbols~$G_k[j_k](\alpha_{k-1}, \alpha_{k})$ to denote the element of the matrix~$\mat{G}_k[j_k]$ in the position $(\alpha_{k-1}, \alpha_{k})$, where $\alpha_{k-1}=1,\ldots,\rank_{k-1}$, $\alpha_k=1,\ldots,\rank_k$.
Equation~\eqref{TT-format} can be equivalently rewritten as the sum of the products of the elements of the cores:
\begin{equation}
    \label{TT-format-sum}
    \tensel{A}(j_1, \ldots, j_d) =\!\!\!\!\!
    \sum_{\alpha_0, \ldots, \alpha_d}\!\!\!\!\! G_1[j_1](\alpha_0, \alpha_1) \ldots G_d[j_d](\alpha_{d-1}, \alpha_d).
\end{equation}

The representation of a tensor~$\tens{A}$ via the explicit enumeration of all its elements requires to store $\prod_{k=1}^d n_k$ numbers compared with $\sum_{k=1}^d n_k \rank_{k-1} \rank_{k}$ numbers if the tensor is stored in the TT-format. Thus, the TT-format is very efficient in terms of memory if the  ranks are small.

An attractive property of the TT-decomposition is the ability to efficiently perform several types of operations on tensors if they are in the TT-format:
basic linear algebra operations, such as the addition of a constant and the multiplication by a constant, the summation and the entrywise product of tensors (the results of these operations are tensors in the TT-format generally with the increased ranks); computation of global characteristics of a tensor, such as the sum of all elements and the Frobenius norm. See~\cite{oseledets2011ttMain} for a detailed description of all the supported operations. 

\subsection{TT-representations for vectors and matrices}
The direct application of the TT-decomposition to a matrix (2-dimensional tensor) coincides with the low-rank matrix format and the direct TT-decomposition of a vector is equivalent to explicitly storing its elements. To be able to efficiently work with large vectors and matrices the TT-format for them is defined in a special manner.
Consider a vector $\vec{b}\in\mathbb{R}^N$, where $N=\prod_{k=1}^dn_k$. We can establish a bijection $\vec{\mu}$ between the coordinate $\ell \in \{1,\ldots,N\}$ of $\vec{b}$ and a $d$-dimensional vector-index $\vec{\mu}(\ell)=(\mu_1(\ell),\ldots,\mu_d(\ell))$ of the corresponding tensor $\tens{B}$, where $\mu_k(\ell) \in \{1,\ldots,n_k\}$. The tensor~$\tens{B}$ is then defined by the corresponding vector elements: $\tensel{B}(\vec{\mu}(\ell))=b_\ell$. Building a TT-representation of $\tens{B}$ allows us to establish a compact format for the vector $\vec{b}$. We refer to it as a \emph{TT-vector}.

Now we define a TT-representation of a matrix~$\mat{W}\in\mathbb{R}^{M\times N}$, where $M=\prod_{k=1}^dm_k$ and $N=\prod_{k=1}^d n_k$. Let bijections $\vec{\nu}(t)=(\nu_1(t),\ldots,\nu_d(t))$ and $\vec{\mu}(\ell)=(\mu_1(\ell),\ldots,\mu_d(\ell))$ map row and column indices $t$ and $\ell$ of the matrix~$\mat{W}$ to the $d$-dimensional vector-indices whose $k$-th dimensions are of length $m_k$ and $n_k$ respectively, $k=1,\ldots,d$. From the matrix~$\mat{W}$ we can form a $d$-dimensional tensor~$\tens{W}$ whose $k$-th dimension is of length $m_k n_k$ and is indexed by the tuple $(\nu_k(t),\mu_k(\ell))$. The tensor $\tens{W}$ can then be converted into the TT-format:
\begin{equation}
\label{eq:matrix-tt-format}
W(t,\ell) = \tensel{W}((\nu_1(t), \mu_1(\ell)),\ldots,(\nu_d(t),\mu_d(\ell))) = \mat{G}_1[\nu_1(t),\mu_1(\ell)] \dots \mat{G}_d[\nu_d(t),\mu_d(\ell)],
\end{equation}
where the matrices~$\mat{G}_k[\nu_k(t),\mu_k(\ell)]$, $k = 1,\dots,d$, serve as the cores with tuple $(\nu_k(t),\mu_k(\ell))$ being an index.
Note that a matrix in the TT-format is not restricted to be square. Although index-vectors~$\vec{\nu}(t)$ and~$\vec{\mu}(\ell)$ are of the same length $d$, the sizes of the domains of the dimensions can vary. We call a matrix in the TT-format a \emph{TT-matrix}.

All operations available for the TT-tensors are applicable to the TT-vectors and the TT-matrices as well (for example one can efficiently sum two TT-matrices and get the result in the TT-format). Additionally, the TT-format allows to efficiently perform the matrix-by-vector (matrix-by-matrix) product. If only one of the operands is in the TT-format, the result would be an explicit vector (matrix); if both operands are in the TT-format, the operation would be even more efficient and the result would be given in the TT-format as well (generally with the increased ranks). For the case of the TT-matrix-by-explicit-vector product $\vec{c}=\mat{W} \vec{b}$, the computational complexity is $O(d \, r^2 \, m \max\{M, N\})$, where $d$ is the number of the cores of the TT-matrix $\mat{W}$, $m = \max_{k=1,\ldots,d} m_k$, $r$ is the maximal rank and $N = \prod_{k=1}^d n_k$ is the length of the vector $\vec{b}$.

The ranks and, correspondingly, the efficiency of the TT-format for a vector (matrix) depend on the choice of the mapping $\vec{\mu}(\ell)$ (mappings $\vec{\nu}(t)$ and $\vec{\mu}(\ell)$) between vector (matrix) elements and the underlying tensor elements. In what follows we use a column-major MATLAB \texttt{reshape} command~\footnote{\url{http://www.mathworks.com/help/matlab/ref/reshape.html}} to form a $d$-dimensional tensor from the data (e.\,g. from a multichannel image), but one can choose a different mapping.

\section{TT-layer \label{sec:TT-layer}}
In this section we introduce the \emph{TT-layer} of a neural network.
In short, the TT-layer is a fully-connected layer with the weight matrix stored in the TT-format. We will refer to a neural network with one or more TT-layers as \emph{TensorNet}.

Fully-connected layers apply a linear transformation to an $N$-dimensional input vector $\vec{x}$:
\begin{equation}
\label{eq:standard-fc}
\vec{y} = \mat{W} \vec{x} + \vec{b},
\end{equation}
where the \emph{weight matrix} $\mat{W} \in \mathbb{R}^{M \times N}$ and the \emph{bias vector} $\vec{b} \in \mathbb{R}^M$ define the transformation.

A \emph{TT-layer} consists in storing the weights~$\mat{W}$ of the fully-connected layer in the TT-format, allowing to use hundreds of thousands (or even millions) of hidden units while having moderate number of parameters. To control the number of parameters one can vary the number of hidden units as well as the TT-ranks of the weight matrix.

A TT-layer transforms a $d$-dimensional tensor $\tens{X}$ (formed from the corresponding vector $\vec{x}$) to the $d$-dimensional tensor $\tens{Y}$ (which correspond to the output vector $\vec{y}$). We assume that the weight matrix $\mat{W}$ is represented in the TT-format with the cores $\mat{G}_k[i_k,j_k]$. The linear transformation~\eqref{eq:standard-fc} of a fully-connected layer can be expressed in the tensor form:
\begin{equation}
\label{eq:TT-layer-output-detailed}
\tensel{Y}(i_1, \ldots, i_d) =
\sum_{j_1, \ldots, j_d}  \!\!\mat{G}_1[i_1, j_1] \dots \mat{G}_d[i_d, j_d]\, \tensel{X}(j_1,\ldots,j_d) + \tensel{B}(i_1,\ldots,i_d).
\end{equation}

Direct application of the TT-matrix-by-vector operation for the Eq.~\eqref{eq:TT-layer-output-detailed} yields the computational complexity of the forward pass $O(d r^2 m \max\{m, n\}^d) = O(d r^2 m \max\{M, N\})$.

\section{Learning \label{sec:learning}}
Neural networks are usually trained with the stochastic gradient descent algorithm where the gradient is computed using the back-propagation procedure~\cite{rumelhart1986}. Back-propagation allows to compute the gradient of a loss-function $L$ with respect to all the parameters of the network. The method starts with the computation of the gradient of $L$ w.r.t. the output of the last layer and proceeds sequentially through the layers in the reversed order while computing the gradient w.r.t. the parameters and the input of the layer making use of the gradients computed earlier. Applied to the fully-connected layers~\eqref{eq:standard-fc} the back-propagation method computes the gradients w.r.t. the input~$\vec{x}$ and the parameters $\mat{W}$ and $\vec{b}$ given the gradients~$\frac{\partial L}{\partial \vec{y}}$ w.r.t to the output~$\vec{y}$:
\begin{equation}
\label{eq:traditional-gradient}
\frac{\partial L}{\partial \vec{x}} = \mat{W}^\intercal \frac{\partial L}{\partial \vec{y}}, ~~~~~\frac{\partial L}{\partial \mat{W}} = \frac{\partial L}{\partial \vec{y}} \vec{x}^\intercal, ~~~~~ \frac{\partial L}{\partial \vec{b}} = \frac{\partial L}{\partial \vec{y}}.
\end{equation}

In what follows we derive the gradients required to use the back-propagation algorithm with the TT-layer. To compute the gradient of the loss function w.r.t. the bias vector $\vec{b}$ and w.r.t. the input vector $\vec{x}$ one can use equations~\eqref{eq:traditional-gradient}. The latter can be applied using the matrix-by-vector product (where the matrix is in the TT-format) with the complexity of $O(d r^2 n \max\{m, n\}^d) = O(d r^2 n \max\{M, N\})$.

To perform a step of stochastic gradient descent one can use equation~\eqref{eq:traditional-gradient} to compute the gradient of the loss function w.r.t. the weight matrix $\mat{W}$, convert the gradient matrix into the TT-format (with the TT-SVD algorithm~\cite{oseledets2011ttMain}) and then add this gradient (multiplied by a step size) to the current estimate of the weight matrix: $\mat{W}_{k+1} = \mat{W}_{k} + \gamma_k \frac{\partial L}{\partial \mat{W}}$. However, the direct computation of $\frac{\partial L}{\partial \mat{W}}$ requires $O(MN)$ memory.
A better way to learn the TensorNet parameters is to compute the gradient of the loss function directly w.r.t. the cores of the TT-representation of $\mat{W}$.

In what follows we use shortened notation for prefix and postfix sequences of indices: $\vec{i}_k^- := (i_1, \dots, i_{k-1})$, $\vec{i}_k^+ := (i_{k+1}, \dots, i_d)$, $\vec{i} = (\vec{i}_k^-, i_k, \vec{i}_k^+)$. We also introduce notations for partial core products:
\begin{equation}
\begin{aligned}
\vec{P}_k^-[\vec{i}_k^-, \vec{j}_k^-] &:= \vec{G}_1[i_1, j_1] \dots \vec{G}_{k-1}[i_{k-1}, j_{k-1}], \\
\vec{P}_k^+[\vec{i}_k^+, \vec{j}_k^+] &:= \vec{G}_{k+1}[i_{k+1}, j_{k+1}] \dots \vec{G}_d[i_d, j_d].
\end{aligned}
\end{equation}
We now rewrite the definition of the TT-layer transformation~\eqref{eq:TT-layer-output-detailed} for any $k = 2, \ldots, d-1$:
\begin{equation}
\label{eq:TT-layer-tensor-form}
\tensel{Y}(\vec{i}) = \tensel{Y}(\vec{i}_k^-, i_k, \vec{i}_k^+) =
\sum_{\vec{j}_k^-, j_k, \vec{j}_k^+}  \vec{P}_k^-[\vec{i}_k^-, \vec{j}_k^-] \mat{G}_k[i_k, j_k] \vec{P}_k^+[\vec{i}_k^+, \vec{j}_k^+] \tensel{X}(\vec{j}_k^-, j_k, \vec{j}_k^+) + \tensel{B}(\vec{i}).
\end{equation}

The gradient of the loss function $L$ w.r.t. to the $k$-th core in the position $[\tilde{i}_k, \tilde{j}_k]$ can be computed using the chain rule:
\begin{equation}
\label{eq:d-L-d-G}
\underbrace{\frac{\partial{L}}{\partial{\vec{G}_k[\tilde{i}_k, \tilde{j}_k]}}}_{\rank_{k-1} \times \rank_{k}} = \sum_{\vec{i}} \frac{\partial{L}}{\partial{\tensel{Y}(\vec{i})}} \frac{\partial{\tensel{Y}}(\vec{i})}{\partial{\vec{G}_k[\tilde{i}_k, \tilde{j}_k]}}.
\end{equation}
Given the gradient matrices $\frac{\partial{\tensel{Y}}(\vec{i})}{\partial{\vec{G}_k[\tilde{i}_k, \tilde{j}_k]}}$ the summation~\eqref{eq:d-L-d-G} can be done explicitly in $O(M \rank_{k-1} \rank_{k})$ time, where $M$ is the length of the output vector $\vec{y}$.

\begin{table}\begin{center}
    \begin{tabular}{ l | l | l }
    Operation & Time & Memory \rule{0pt}{1.0\normalbaselineskip} \\ \hline
    FC forward pass & $O(M N)$ & $O(M N)$ \rule{0pt}{1.0\normalbaselineskip}\\ 
    TT forward pass & $O(d r^2 m \max\{M, N\})$ & $O(r \max\{M, N\})$ \\ 
    FC backward pass & $O(M N)$ & $O(M N)$ \\ 
    TT backward pass & $O(d^2 \rank^4 m \max\{M, N\})$ & $O(\rank^3 \max\{M, N\})$ \\[-0.25cm] 
    \end{tabular}
    \end{center}
    \caption{Comparison of the asymptotic complexity and memory usage of an $M \times N$ TT-layer and an $M \times N$ fully-connected layer (FC). The input and output tensor shapes are $m_1 \times \ldots \times m_d$ and $n_1 \times \ldots \times n_d$ respectively ($m = \max_{k = 1 \ldots d} m_k$) and $\rank$ is the maximal TT-rank.\label{tbl:complexity-comparison}\vspace{-0.4cm}}
\end{table}

We now show how to compute the matrix $\frac{\partial{\tensel{Y}}(\vec{i})}{\partial{\vec{G}_k[\tilde{i}_k, \tilde{j}_k]}}$ for any values of the core index $k \in \{1, \dots, d\}$ and $\tilde{i}_k \in \{1, \dots, m_k\}$, $\tilde{j}_k \in \{1, \dots, n_k\}$.
For any $\vec{i} = (i_1, \dots, i_d)$ such that $i_k \neq \tilde{i}_k$ the value of $\tensel{Y}(\vec{i})$ doesn't depend on the elements of $\mat{G}_k[\tilde{i}_k, \tilde{j}_k]$ making the corresponding gradient $\frac{\partial \tensel{Y}(\vec{i})}{\partial \mat{G}_k[\tilde{i}_k, \tilde{j}_k]}$ equal zero. Similarly, any summand in the Eq.~\eqref{eq:TT-layer-tensor-form} such that $j_k \neq \tilde{j}_k$ doesn't affect the gradient $\frac{\partial{\tensel{Y}}(\vec{i})}{\partial{\vec{G}_k[\tilde{i}_k, \tilde{j}_k]}}$. These observations allow us to consider only $i_k = \tilde{i}_k$ and $j_k = \tilde{j}_k$.

$\tensel{Y}(\vec{i}_k^-, \tilde{i}_k, \vec{i}_k^+)$ is a linear function of the core $\mat{G}_k[\tilde{i}_k, \tilde{j}_k]$ and its gradient equals the following expression:
\begin{align}
\label{eq:jacobian}
\frac{\partial{\tensel{Y}(\vec{i}_k^-, \tilde{i}_k, \vec{i}_k^+)}}{\partial{\vec{G}_k[\tilde{i}_k, \tilde{j}_k]}} = \sum_{\vec{j}_k^-, \vec{j}_k^+} \underbrace{\left ( \vec{P}_k^-[\vec{i}_k^-, \vec{j}_k^-] \right )^\intercal }_{\rank_{k-1} \times 1}  \underbrace{\left (\vec{P}_k^+[\vec{i}_k^+, \vec{j}_k^+] \right )^\intercal}_{1 \times \rank_{k}} \tensel{X}(\vec{j}_k^-, \tilde{j}_k, \vec{j}_k^+).
\end{align}

We denote the partial sum vector as $\vec{R}_{k}[\vec{j}_{k}^-, \tilde{j}_k, \vec{i}_k^+] \in \mathbb{R}^{\rank_{k}}$:
\begin{align*}
 \vec{R}_{k}[j_1, \dots, j_{k-1}, \tilde{j}_k, i_{k+1}, \dots, i_d]  =  \vec{R}_{k}[\vec{j}_{k}^-, \tilde{j}_k, \vec{i}_k^+]  = \sum_{\vec{j}_k^+} \vec{P}_k^+[\vec{i}_k^+, \vec{j}_k^+]  ~ \tensel{X}(\vec{j}_k^-, \tilde{j}_k, \vec{j}_k^+).
\end{align*}
Vectors $\vec{R}_{k}[\vec{j}_{k}^-, \tilde{j}_k, \vec{i}_k^+]$ for all the possible values of $k$, $\vec{j}_{k}^-$, $\tilde{j}_k$ and $\vec{i}_k^+$ can be computed via dynamic programming (by pushing sums w.r.t. each $j_{k+1}, \ldots, j_d$ inside the equation and summing out one index at a time) in $O(d r^2 m \max\{M, N\})$. Substituting these vectors into~\eqref{eq:jacobian} and using (again) dynamic programming yields us all the necesary matrices for summation~\eqref{eq:d-L-d-G}. The overall computational complexity of the backward pass is $O(d^2 \rank^4 m \max\{M, N\})$.



The presented algorithm reduces to a sequence of matrix-by-matrix products and permutations of dimensions and thus can be accelerated on a GPU device.

\section{Experiments \label{sec:experiments}\vspace{-0.1cm}}

\subsection{Parameters of the TT-layer \label{sec:mnist-shapes}\vspace{-0.1cm}}

In this experiment we investigate the properties of the TT-layer and compare different strategies for setting its parameters: dimensions of the tensors representing the input/output of the layer and the TT-ranks of the compressed weight matrix. We run the experiment on the MNIST dataset~\cite{lecun1998mnist} for the task of handwritten-digit recognition.
As a baseline we use a neural network with two fully-connected layers ($1024$ hidden units) and rectified linear unit (ReLU) achieving  $1.9\%$ error on the test set.
For more reshaping options we resize the original $28 \times 28$ images to $32 \times 32$.

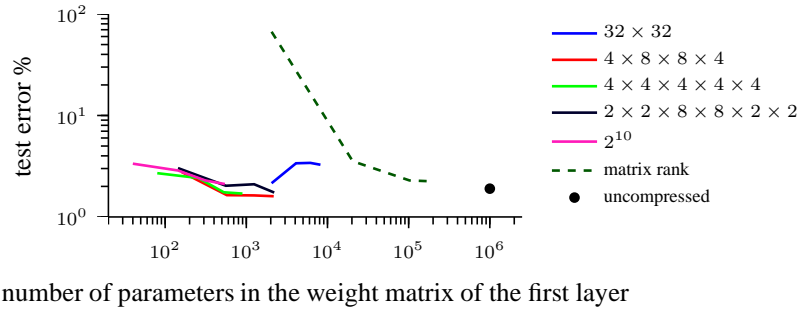
\begin{figure}[t]
  \begin{center}
  \setlength\figureheight{2,7cm}
  \setlength\figurewidth{5.5cm}
%
%
%
\definecolor{mycolor1}{rgb}{0.00000,0.00000,0.17241}%
\definecolor{mycolor2}{rgb}{1.00000,0.10345,0.72414}%
\definecolor{mycolor3}{rgb}{0.00000,0.34483,0.00000}%
\begin{tikzpicture}

\begin{axis}[%
width=\figurewidth,
height=\figureheight,
scale only axis,
xmode=log,
xmin=20,
xmax=2500000,
xminorticks=true,
xlabel={number of parameters in the weight matrix of the first layer},
ymode=log,
ymin=1,
ymax=100,
yminorticks=true,
ylabel={test error \%},
axis x line*=bottom,
axis y line*=left,
legend style={draw=none,fill=white,legend cell align=left,at={(1.7,1)},anchor=north east}
]
\addplot [color=blue,solid]
  table[row sep=crcr]{2048  2.15\\
4096    3.38\\
6144    3.41\\
8192    3.27\\
};
\addlegendentry{$32 \times 32$};

\addplot [color=red,solid]
  table[row sep=crcr]{160   2.77\\
576 1.64\\
1248    1.63\\
2176    1.6\\
};
\addlegendentry{$4 \times 8 \times 8 \times 4$};

\addplot [color=green,solid]
  table[row sep=crcr]{80    2.7\\
256 2.42\\
528 1.75\\
896 1.7\\
};
\addlegendentry{$4 \times 4 \times 4 \times 4 \times 4$};

\addplot [color=mycolor1,solid]
  table[row sep=crcr]{144   3.03\\
560 2.03\\
1248    2.1\\
2208    1.74\\
};
\addlegendentry{$2 \times 2 \times 8 \times 8 \times 2 \times 2$};

\addplot [color=mycolor2,solid]
  table[row sep=crcr]{40    3.35\\
144 2.87\\
312 2.29\\
544 2.11\\
};
\addlegendentry{$2^{10}$};

\addplot [color=mycolor3,dashed]
  table[row sep=crcr]{2048  66.89\\
20480   3.54\\
102400  2.29\\
204800  2.24\\
};
\addlegendentry{matrix rank};

\addplot [only marks,color=black,solid,mark=*,mark options={solid},mark size=1.5pt]
  table[row sep=crcr]{1000000 1.9\\
};
\addlegendentry{uncompressed};

\end{axis}
\end{tikzpicture}%
  \end{center}
  \vspace{-0.5cm}
  \caption{The experiment on the MNIST dataset. We use a two-layered neural network and substitute the first $1024 \times 1024$ fully-connected layer with the TT-layer (solid lines) and with the matrix rank decomposition based layer (dashed line). The solid lines of different colors correspond to different ways of reshaping the input and output vectors to tensors (the shapes are reported in the legend). To obtain the points of the plots we vary the maximal TT-rank or the matrix rank. \label{fig:mnist-shape}\vspace{-0.5cm}}
\end{figure}

We train several networks differing in the parameters of the single TT-layer. The networks contain the following layers: the TT-layer with weight matrix of size $1024 \times 1024$, ReLU, the fully-connected layer with the weight matrix of size $1024 \times 10$. We test different ways of reshaping the input/output tensors and try different ranks of the TT-layer. As a simple compression baseline in the place of the TT-layer we use the fully-connected layer such that the rank of the weight matrix is bounded (implemented as follows: the two consecutive fully-connected layers with weight matrices of sizes $1024 \times \rank$ and $\rank \times 1024$, where $\rank$ controls the matrix rank and the compression factor).
The results of the experiment are shown in Figure~\ref{fig:mnist-shape}.
We conclude that the TT-ranks provide much better flexibility than the matrix rank when applied at the same compression level. In addition, we observe that the TT-layers with too small number of values for each tensor dimension and with too few dimensions perform worse than their more balanced counterparts.

\paragraph{Comparison with HashedNet~\cite{chen2015compressing}.} We consider a two-layered neural network with $1024$ hidden units and replace both fully-connected layers by the TT-layers. By setting all the TT-ranks in the network to $8$ we achieved the test error of $1.6\%$ with $12\,602$ parameters in total and by setting all the TT-ranks to $6$ the test error of $1.9\%$ with $7\,698$ parameters. \citet{chen2015compressing} report results on the same architecture. By tying random subsets of weights they compressed the network by the factor of $64$ to the  $12\,720$ parameters in total with the test error equal $2.79\%$.

\subsection{CIFAR-10}
CIFAR-10 dataset~\cite{krizhevsky2009Cifar} consists of $32 \times 32$ 3-channel images assigned to  10 different classes: airplane, automobile, bird, cat, deer, dog, frog, horse, ship, truck. The dataset contains 50000 train and 10000 test images. Following~\cite{goodfellow2013maxout} we preprocess the images by subtracting the mean and performing global contrast normalization and ZCA whitening.

As a baseline we use the CIFAR-10 Quick~\cite{snoek2012cifarQuick} CNN, which consists of convolutional, pooling and non-linearity layers followed by two fully-connected layers of sizes $1024 \times 64$ and $64 \times 10$. We fix the convolutional part of the network and substitute the fully-connected part by a $1024 \times N$ TT-layer followed by ReLU and by a $N \times 10$ fully-connected layer. With $N = 3125$ hidden units (contrary to $64$ in the original network) we achieve the test error of $23.13\%$ without fine-tuning which is slightly better than the test error of the baseline ($23.25\%$). The TT-layer treated input and output vectors as $4 \times 4 \times 4 \times 4 \times 4$ and $5 \times 5 \times 5 \times 5 \times 5$ tensors respectively. All the TT-ranks equal $8$, making the number of the parameters in the TT-layer equal $4\,160$. The compression rate of the TensorNet compared with the baseline w.r.t. all the parameters is $1.24$.
In addition, substituting the both fully-connected layers by the TT-layers yields the test error of $24.39\%$ and reduces the number of parameters of the fully-connected layer matrices by the factor of $11.9$ and the total parameter number by the factor of $1.7$.

For comparison, in~\cite{Denil2013predicting} the fully-connected layers in a CIFAR-10 CNN were compressed by the factor of at most $4.7$ times with the loss of about $2\%$ in accuracy.

\subsubsection{Wide and shallow network}
With sufficient amount of hidden units, even a neural network with two fully-connected layers and sigmoid non-linearity can approximate any decision boundary~\cite{cybenko1989universalApproximator}. Traditionally, very wide shallow networks are not considered because of high computational and memory demands and the over-fitting risk. TensorNet can potentially address both issues. We use a three-layered TensorNet of the following architecture: the TT-layer with the weight matrix of size $3\,072 \times 262\,144$, ReLU, the TT-layer with the weight matrix of size $262\,144 \times 4\,096$, ReLU, the fully-connected layer with the weight matrix of size $4\,096 \times 10$. We report the test error of $31.47\%$ which is (to the best of our knowledge) the best result achieved by a non-convolutional neural network.

\begin{table}
    \begin{center}
    \begin{tabular}{ l@{\;}|@{\;}c@{\;}|@{\;}c@{\;}|@{\;}c@{\;}|@{\;}c@{\;}|@{\;}c@{\;}|@{\;}c@{\;}|@{\;}c@{\;} }
    Architecture &
    \parbox{1.4cm}{TT-layers \\ compr.} &
    \parbox{1.0cm}{vgg-16 \\ compr.} &
    \parbox{1.0cm}{vgg-19 \\ compr.} &
    \parbox{1.0cm}{vgg-16 \\ top 1} &
    \parbox{1.0cm}{vgg-16 \\ top 5} &
    \parbox{1.0cm}{vgg-19 \\ top 1} &
    \parbox{1.0cm}{vgg-19 \\ top 5} \rule{0pt}{1.0\normalbaselineskip} \\ \hline
    FC FC FC \rule{0pt}{1.0\normalbaselineskip}     & $1$ &  $1$ & $1$ & $30.9$ &  $11.2$ & $29.0$ & $10.1$  \\ 
    TT4 FC FC    & $50\,972$ & $3.9$ & $3.5$ & $31.2$ & $11.2$ & $29.8$ & $10.4$ \\ 
    TT2 FC FC    & $194\,622$ & $3.9$ & $3.5$  & $31.5$ & $11.5$ & $30.4$ & $10.9$ \\ 
    TT1 FC FC    & $713\,614$ & $3.9$ & $3.5$  & $33.3$ & $12.8$ & $31.9$ & $11.8$ \\ 
    TT4 TT4 FC    & $37\,732$ & $7.4$ & $6$ & $32.2$ & $12.3$ & $31.6$ & $11.7$ \\ 
    MR1 FC FC    & $3\,521$ & $3.9$ & $3.5$ & $99.5$ & $97.6$ & $99.8$ & $99$ \\ 
    MR5 FC FC    & $704$ & $3.9$ & $3.5$ & $81.7$ & $53.9$ & $79.1$ & $52.4$ \\ 
    MR50 FC FC    & $70$ & $3.7$ & $3.4$ & $36.7$ & $14.9$ & $34.5$ & $15.8$ \\[-0.3cm] 
    \end{tabular}
    \end{center}
    \caption{Substituting the fully-connected layers with the TT-layers in vgg-16 and vgg-19 networks on the ImageNet dataset. FC stands for a fully-connected layer; TT$\square$ stands for a TT-layer with all the TT-ranks equal ``$\square$''; MR$\square$ stands for a fully-connected layer with the matrix rank restricted to ``$\square$''. We report the compression rate of the TT-layers matrices and of the whole network in the second, third and fourth columns. \label{tbl:imagenet-vgg-layer}\vspace{-0.5cm}}
\end{table}

\subsection{ImageNet}
In this experiment we evaluate the TT-layers on a large scale task. We consider the 1000-class ImageNet ILSVRC-2012 dataset~\cite{Russakovsky2015ImageNet}, which consist of 1.2 million training images and 50 000 validation images. We use deep the CNNs vgg-16 and vgg-19~\cite{simonyan15} as the reference models\footnote{After we had started to experiment on the vgg-16 network the vgg-* networks have been improved by the authors. Thus, we report the results on a slightly outdated version of vgg-16 and the up-to-date version of vgg-19.}\!\!. Both networks consist of the two parts: the convolutional and the fully-connected parts. In the both networks the second part consist of $3$ fully-connected layers with weight matrices of sizes $25088 \times 4096$, $4096 \times 4096$ and $4096 \times 1000$.

In each network we substitute the first fully-connected layer with the TT-layer.
To do this we reshape the $25088$-dimensional input vectors to the tensors of the size $2 \times 7 \times 8 \times 8 \times 7 \times 4$ and the $4096$-dimensional output vectors to the tensors of the size $4 \times 4 \times 4 \times 4 \times 4 \times 4$.
The remaining fully-connected layers are initialized randomly.
The parameters of the convolutional parts are kept fixed as trained by~\citet{simonyan15}.
We train the TT-layer and the fully-connected layers on the training set.
In Table~\ref{tbl:imagenet-vgg-layer} we vary the ranks of the TT-layer and report the compression factor of the TT-layers (vs. the original fully-connected layer), the resulting compression factor of the whole network, and the top~1 and top~5 errors on the validation set.
In addition, we substitute the second fully-connected layer with the TT-layer.
As a baseline compression method we constrain the matrix rank of the weight matrix of the first fully-connected layer using the approach of~\cite{Jimmy2014lowRankSNN}.

In Table~\ref{tbl:imagenet-vgg-layer} we observe that the TT-layer in the best case manages to reduce the number of the parameters in the matrix~$\mat{W}$ of the largest fully-connected layer by a factor of $194\,622$ (from $25088 \times 4096$ parameters to $528$) while increasing the top 5 error from $11.2$ to $11.5$. The compression factor of the whole network remains at the level of~$3.9$ because the TT-layer stops being the storage bottleneck. By compressing the largest of the remaining layers the compression factor goes up to~$7.4$. The baseline method when providing similar compression rates significantly increases the error.

For comparison, consider the results of~\cite{yang2014deep} obtained for the compression of the fully-connected layers of the Krizhevsky-type network~\cite{Krizhevsky2012AlexNet} with the Fastfood method. The model achieves compression factors of 2-3 without decreasing the network error.


\begin{table}
    \begin{center}
    \begin{tabular}{l| l l}
    Type & 1 im. time (ms) & 100 im. time (ms) \rule{0pt}{1.0\normalbaselineskip}\\ \hline
    CPU fully-connected layer   & $16.1$ & $97.2$ \rule{0pt}{1.0\normalbaselineskip}\\
    CPU TT-layer   & $1.2$ & $94.7$\\
    GPU fully-connected layer   & $2.7$ & $33$\\
    GPU TT-layer   & $1.9$ & $12.9$\\[-0.3cm]
    \end{tabular}
    \end{center}
    \caption{Inference time for a $25088 \times 4096$ fully-connected layer and its corresponding TT-layer with all the TT-ranks equal $4$. The memory usage for feeding forward one image is $392$MB for the fully-connected layer and $0.766$MB for the TT-layer. \label{tbl:imegenet-inference}\vspace{-0.5cm}}
\end{table}

\subsection{Implementation details}
In all experiments we use our MATLAB extension\footnote{\url{https://github.com/Bihaqo/TensorNet}} of the MatConvNet framework\footnote{\url{http://www.vlfeat.org/matconvnet/}}~\cite{matconvnet}. For the operations related to the TT-format we use the TT-Toolbox\footnote{\url{https://github.com/oseledets/TT-Toolbox}}
implemented in MATLAB as well. The experiments were performed on a computer with a quad-core Intel Core i5-4460 CPU, 16 GB RAM and a single NVidia Geforce GTX 980 GPU.
We report the running times and the memory usage at the forward pass of the TT-layer and the baseline fully-connected layer in Table~\ref{tbl:imegenet-inference}.

We train all the networks with stochastic gradient descent with momentum (coefficient $0.9$). We initialize all the parameters of the TT- and fully-connected layers with a Gaussian noise and put L$2$-regularization (weight~$0.0005$) on them.


\section{Discussion and future work \label{sec:discussion}}
Recent studies indicate high redundancy in the current neural network parametrization. To exploit this redundancy we propose to use the TT-decomposition framework on the weight matrix of a fully-connected layer and to use the cores of the decomposition as the parameters of the layer. This allows us to train the fully-connected layers compressed by up to $200\,000\times$ compared with the explicit parametrization without significant error increase. Our experiments show that it is possible to capture complex dependencies within the data by using much more compact representations. On the other hand it becomes possible to use much wider layers than was available before and the preliminary experiments on the CIFAR-10 dataset show that wide and shallow TensorNets achieve promising results (setting new state-of-the-art for non-convolutional neural networks).

Another appealing property of the TT-layer is faster inference time (compared with the corresponding fully-connected layer). All in all a wide and shallow TensorNet can become a time and memory efficient model to use in real time applications and on mobile devices.

The main limiting factor for an $M \times N$ fully-connected layer size is its parameters number $MN$. The limiting factor for an $M \times N$ TT-layer is the maximal linear size $\max\{M, N\}$. As a future work we plan to consider the inputs and outputs of layers in the TT-format thus completely eliminating the dependency on $M$ and $N$ and allowing billions of hidden units in a TT-layer.

\paragraph{Acknowledgements.}
We would like to thank Ivan Oseledets for valuable discussions.
A.\,Novikov, D.\,Podoprikhin, D.\,Vetrov were supported by RFBR project No.\,15-31-20596 (mol-a-ved) and by Microsoft: Moscow State University Joint Research Center (RPD 1053945).
A.\,Osokin was supported by the MSR-INRIA Joint Center. The results of the tensor toolbox application (in Sec.~\ref{sec:experiments}) are supported by Russian Science Foundation No.\,14-11-00659.

\nocite{matconvnet}
\small{
\bibliography{tensor-net}
\bibliographystyle{IEEEtranSN}
}
\end{document}